\documentclass{article}

\PassOptionsToPackage{numbers, compress}{natbib}

\usepackage[preprint]{neurips_2022}

\usepackage[utf8]{inputenc} 
\usepackage[T1]{fontenc}    
\usepackage{url}            
\usepackage{booktabs}       
\usepackage{amsfonts}       
\usepackage{nicefrac}       
\usepackage{microtype}      
\usepackage{xcolor}         

\usepackage{epsfig}
\usepackage{graphicx}
\usepackage{amsmath}
\usepackage{amssymb}
\usepackage{tabulary,multirow,overpic}
\usepackage{dsfont}
\usepackage[position=top]{subfig}
\usepackage{sidecap}
\usepackage{wrapfig}
\usepackage{makecell}
\usepackage[export]{adjustbox}
\setcitestyle{square} 
\usepackage{pifont}

\usepackage{marvosym}
\definecolor{citecolor}{RGB}{34,139,34}
\usepackage[colorlinks,citecolor=citecolor]{hyperref}

\usepackage{xspace}

\makeatletter
\DeclareRobustCommand\onedot{\futurelet\@let@token\@onedot}
\def\@onedot{\ifx\@let@token.\else.\null\fi\xspace}

\def\eg{\emph{e.g}\onedot} 
\def\ie{\emph{i.e}\onedot} 
 
 \def\vs{\emph{vs}\onedot}

\usepackage{extarrows}
\usepackage{color,colortbl}
\definecolor{LightCyan}{rgb}{0.88,1,1}

\title{Towards Model Generalization for Monocular \\ 3D Object Detection}

\author{%
Zhenyu Li$^{1}$ \quad Zehui Chen$^{2}$ \quad Ang Li$^{3}$ \quad Liangji Fang$^{3}$ \\ \textbf{Qinhong Jiang$^{3}$} \quad  \textbf{Xianming Liu$^{1}$} \quad \textbf{Junjun Jiang$^{1}$\textsuperscript{\Letter}} \\
$^{1}$Harbin Institute of Technology \\ $^{2}$University of Science and Technology of China \\ $^{3}$SenseTime Research
}

\begin{document}

\maketitle

\begin{abstract}
  Monocular 3D object detection (Mono3D) has achieved tremendous improvements with emerging large-scale autonomous driving datasets and the rapid development of deep learning techniques. However, caused by severe domain gaps (\eg, the field of view (FOV), pixel size, and object size among datasets), Mono3D detectors have difficulty in generalization, leading to drastic performance degradation on unseen domains. To solve these issues, we combine the position-invariant transform and multi-scale training with the pixel-size depth strategy to construct an effective unified \textit{camera-generalized paradigm} (CGP). It fully considers discrepancies in the FOV and pixel size of images captured by different cameras. Moreover, we further investigate the obstacle in quantitative metrics when cross-dataset inference through an exhaustive systematic study. We discern that the size bias of prediction leads to a colossal failure. Hence, we propose the 2D-3D \textit{geometry-consistent object scaling} strategy (GCOS) to bridge the gap via an instance-level augment. Our method called \textbf{DGMono3D} achieves remarkable performance on all evaluated datasets and surpasses the SoTA unsupervised domain adaptation scheme even without utilizing data on the target domain.
\end{abstract}
\section{Introduction}

3D object detection is a critical component for many computer vision applications such as autonomous driving, robot navigation, and virtual reality, to name a few, aiming to categorize and localize objects in 3D space. Previous methods have achieved engaging performance based on the accurate spatial information from multiple sensors, such as LiDAR-scanned point clouds~\cite{zhou2018voxelnet,lang2019pointpillars,shi2020pvrcnn} or stereo images~\cite{chen2020dsgn,li2019stereo,sun2020disp}. With the rapid development of intelligent systems, monocular 3D object detection (Mono3D) from single images has drawn increasing research attention due to the potential prospects of reduced cost and increased modular redundancy. Driven by deep neural networks and large-scale human-annotated datasets~\cite{kesten2019lyft,caesar2020nusc,geiger2012kitti}, this field has obtained remarkable advancements~\cite{zhang2022monodetr,huang2022monodtr,wang2021fcos3d,park2021dd3d,liu2020smoke,wang2019pseudo,chen2016monocular}.

However, trained on one specific domain (\ie, source domain), Mono3D detectors cannot generalize well to a novel test domain (\ie, target domain) due to inevitable domain shifts arising from geographical locations, imaging processes, and object characteristics, which leads to a plight in deploying models for practical applications~\cite{li2022stmono3d}. Though collecting and training with more data from different domains could alleviate this issue, unfortunately, it might be infeasible given diverse real-world scenarios and expensive annotation costs~\cite{yang2021st3d,li2022stmono3d}. Hence, more generalized Mono3D detectors are highly demanded. In this paper, we aim to achieve \textit{single-domain generalization} for monocular 3D object detection, which is more challenging and practical compared with the unsupervised domain adaptation (UDA)~\cite{li2022stmono3d} since the data in the target domain is sometimes inaccessible and we can only obtain few statistics during the training stage.

STMono3D~\cite{li2022stmono3d} makes the first effort for Mono3D UDA. In terms of all discrepancies among datasets (\ie, domain gaps), \cite{li2022stmono3d} conducts a detailed investigation and indicates that the geometry misalignment caused by different camera devices (\ie, intrinsic parameters) will lead to the severe depth-shift phenomenon and hampers the cross-domain inference for the most part. Hence, \cite{li2022stmono3d} proposes a geometry-aligned multi-scale training strategy and adopts the pixel-size depth to make detectors camera-aware. Nevertheless, the detectors still suffer from an imaging misalignment caused by different camera fields of view (FOV). It cannot be awared by models through the proposed strategy~\cite{li2022stmono3d} since the FOV is invariant in resizing images, leading to a sub-optimal solution for model generalization. To bridge this gap, we propose to seamlessly combine the position-invariant transform (PIT)~\cite{gu2021pit}, and multi-scale training with the pixel-size depth strategy~\cite{li2022stmono3d,park2021dd3d,chen2022graphdetr3d} to construct a unified \textit{camera-generalized paradigm} (CGP). It ensures the geometry consistency (\ie, fully considers the discrepancies of FOV and pixel size) when 3D detectors infer on images captured by different cameras and avoids unacceptable overhead costs during the training stage.

While the effective paradigm achieves camera generalization and yields competitive model performance on loose quantitative metrics (\eg, $\text{AP}_{3D}$, $\text{IoU} > 0.5$), we observe the trained detectors cannot obtain satisfactory results (\ie, \textit{average precision drops to zero}) on relatively strict metrics (\eg, $\text{AP}_{3D}$, $\text{IoU} > 0.7$). Moreover, even the STMono3D~\cite{li2022stmono3d} that utilizes images in the target domain for self-training suffers from a similar dilemma. Unlike 2D bounding boxes having a large variety of sizes, depending on the distance of the object from the camera, the size of 3D bounding boxes is more consistent in the same dataset, regardless of the relative location to the camera~\cite{luo2021multialign}. Hence, Mono3D detectors tend to overfit a narrow and dataset-specific distribution of object size from the source domain, which is consistent with observations in LiDAR-based 3D detectors~\cite{wang2020train,luo2021multialign} and leads to the size bias of prediction (Fig.~\ref{fig::statistics}). To better analyze the influence of such a bias on quantitative metrics, we replace the predicted dimensions of object sizes with ground-truth step by step. As shown in Fig.~\ref{fig::upperbound}\textcolor[rgb]{1,0,0}{b}, it is the size bias that leads to the severe degradation of strict metrics. To alleviate this issue, we propose the 2D-3D \textit{geometry-consistent object scaling} strategy (GCOS) that simultaneously scales objects in the 2D image and 3D spatial space and maintains the 2D-3D geometry consistency of objects. This strategy emancipates data in the target domain during the training stage, achieving more generalized Mono3D detectors and effective training pipelines.

In summary, the major contributions of this work are as follows: First, we combine the position-invariant transform~\cite{gu2021pit} and multi-scale training with the pixel-size depth strategy~\cite{li2022stmono3d,park2021dd3d,chen2022graphdetr3d} to construct a unified paradigm for camera-generalized Mono3D detectors, which comprehensively considers FOV and pixel-size discrepancies among domains. Second, we investigate underlying reasons behind the degradation of strict quantitative metrics. Accordingly, we propose the 2D-3D GCOS strategy to alleviate this issue in a data augment manner and boost the model generalization. Third, we conduct extensive experiments on various 3D object detection datasets: KITTI~\cite{geiger2012kitti}, NuSenses~\cite{caesar2020nusc}, and Lyft~\cite{kesten2019lyft}. Our method for \textbf{D}omain \textbf{G}eneralized \textbf{Mono}cular \textbf{3D} detection, named \textbf{DGMono3D}, obtains engaging results. Without training in the target domain, models achieve competitive and even better performance compared with the UDA method~\cite{li2022stmono3d}, demonstrating the effectiveness of DGMono3D.

\section{Related Work}

\textbf{Monocular 3D object detection} has drawn much more attention in recent years~\cite{chen20153d,xu2018multifusion,mousavian20173d,roddick2018orthographic,weng2019monocular,brazil2019m3drpn,wang2022detr3d,wang2021fcos3d,park2021dd3d,wang2022pgd}. Because of the lack in spatial information, earlier work adopts auxiliary depth networks~\cite{chen20153d,xu2018multifusion} or 2D object detectors~\cite{mousavian20173d} to support 3D detection. Another line of study attempts to lift up RGB images into 3D representations, such as OFTNet~\cite{roddick2018orthographic} and Pseudo-Lidar~\cite{weng2019monocular}. To avoid the dependency on sub-networks, recent methods propose to design the framework in an end-to-end manner like 2D detection~\cite{brazil2019m3drpn,liu2020smoke,wang2022detr3d,park2021dd3d}. In this paper, we conduct experiments based on FCOS3D~\cite{wang2021fcos3d}, a neat and representative Mono3D detector that keeps the well-developed designs for 2D detection and is adapted to Mono3D with only fundamental designs for specific 3D targets.

\textbf{Domain adaptation} aims to generalize the model trained on source domains to target domains~\cite{wang2018dasurvey}. Based on whether labels are available on target domains, methods can be divided into supervised and unsupervised, respectively~\cite{wang2018dasurvey}. In detection, most domain adaptation approaches are designed for 2D detectors~\cite{hsu2020progressive,chen2018domain,khodabandeh2019robust,kim2019diversify}, while direct adoption of these techniques to Mono3D may not work well due to the distinct characteristics of targets in the 3D spatial coordinate~\cite{li2022stmono3d}. For domain adaptation methods on LiDAR-based 3D detection~\cite{yang2021st3d,yang2021st3d++,luo2021multialign,zhang2021srdan}, the fundamental differences in data structures and network architectures render these approaches not readily applicable to Mono3D.

STMono3D~\cite{li2022stmono3d} proposes the first UDA paradigm for Mono3D via self-training. While they adopt the multi-scale training with pixel-size depth strategy to overcome the discrepancy of camera parameters when cross-dataset inference, FOV that is invariant in scaling image is neglected and hampers the model performance. Moreover, their methods still suffer from drastic degradation on strict quantitative metrics, as shown in Fig.~\ref{fig::upperbound}. In this paper, we fully consider the pixel size and FOV, conducting the \textit{camera-genralized paradigm} (CGP). Subsequently, we do detailed system analysis and design the 2D-3D \textit{geometry-consistent object scale} strategy (GCOS) to handle the degradation on strict quantitative metrics. Our methods aim to achieve single-domain generalization~\cite{qiao2020learning,wang2021learning,li2021progressive}, and \textit{do not} need images on target domains, which are hard to obtain in practice.

\textbf{Cut-and-paste for detection} is a representative kind of data augment strategy applied to regions that contain objects of an image~\cite{dvornik2018modeling,fang2019instaboost,dwibedi2017cut}. It is also commonly utilized in LiDAR-based 3D detection to process points of objects and is crucial for detector performance~\cite{liu2020tanet,yan2018second,yang2019std}. Besides, there are several explorations for multi-modal 3D detection~\cite{zhang2020moca}, but absent for Mono3D~\cite{wang2021fcos3d,wang2022pgd,wang2022detr3d}. In this paper, we design the GCOS in a cut-and-paste manner. Instead of saving image crops to build a ground-truth database and pasting them randomly during training~\cite{yan2018second,zhang2020moca}, we directly operate objects of current images in an online manner. Moreover, our method also resizes objects and keeps the consistency of 2D-3D correspondence to diminish the size gap on cross-dataset inference.
\begin{figure}[t]
    \centering
    \footnotesize
    \setlength{\tabcolsep}{10pt}
    \scalebox{0.98}{%
    \begin{tabular}{cc}
        \multicolumn{2}{c}{\includegraphics[width=0.96\linewidth]{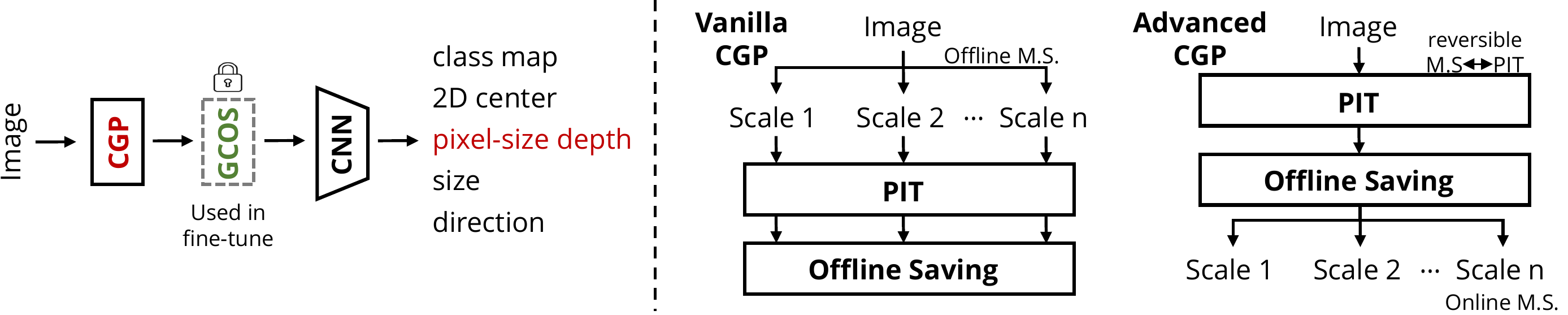}}\\
        \hspace{0.12\linewidth}\scriptsize{(a) Overview} & 
        \hspace{0.16\linewidth}\scriptsize{(b) Camera-Generalized Paradigm} \\
    \end{tabular}}
    \caption{(a) An overview of our proposed DGMono3D, containing a \textit{camera-generalized paradigm} (CGP, \textcolor[rgb]{0.8,0,0}{red}) and a \textit{geometry-consistent object scaling} strategy (GCOS, \textcolor[rgb]{0,0.5,0}{green}). (b) Based on the reversibility of the position-invariant transform (PIT) and the multi-scale strategy (MS), we evolve an advanced CGP that avoids infeasible offline redundancy.} 
    \label{fig::overall&pit}
\end{figure}

\section{DGMono3D}

In this section, we first formulate the single-domain generalization task on Mono3D in Sec.~\ref{sec::subsec::pd} and then present the overall pipeline of DGMono3D in Sec.~\ref{sec::subsec::overview}. Subsequently, we introduce our key contributions in detail, including the camera-generalized paradigm in Sec.~\ref{sec::subsec::cgp} and the 2D-3D geometry-consistent object scale strategy in Sec.~\ref{sec::subsec::gcos}.

\subsection{Problem Definition}
\label{sec::subsec::pd}
Under the single-domain generalization setting, we own labeled images from a \textit{single-source} domain $\mathcal{D}_S=\{x_s^i,y_s^i,K_s^i\}_{i=1}^{N_S}$, and are \textit{inaccessible} to the target domain data $\mathcal{D}_T=\{x_t^i,K_t^i\}_{i=1}^{N_T}$, where $N_s$ and $N_t$ are the number of samples from the source and target domains, respectively. Each 2D image $x^i$ is paired with a camera parameter $K^i$ that associates points in 3D space and 2D image plane while $y_s^i$ denotes the label of the corresponding training sample from the source domain. Label $y$ is in the form of object class $k$, location $(c_x, c_y, c_z)$, size in each dimension $(d_x, d_y, d_z)$, and orientation $\theta$. We aim to train models with $\mathcal{D}_S$ and avoid performance degradation when inferring in any other target domain $\mathcal{D}_T$. Images $\{x_t\}$ and camera parameters $\{K_t\}$ to re-project predictions to 3D locations are available during inference in target domains.

\begin{figure}[t]
    \centering
    \footnotesize
    \setlength{\tabcolsep}{10pt}
    \scalebox{0.95}{%
    \begin{tabular}{ccc}
        \includegraphics[width=0.3\linewidth]{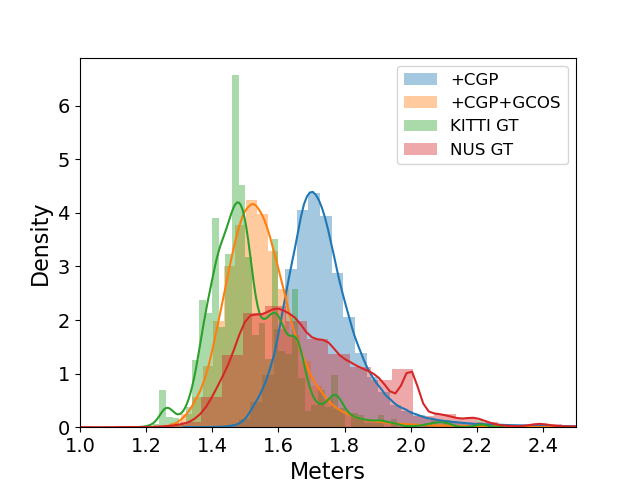}&
        \includegraphics[width=0.3\linewidth]{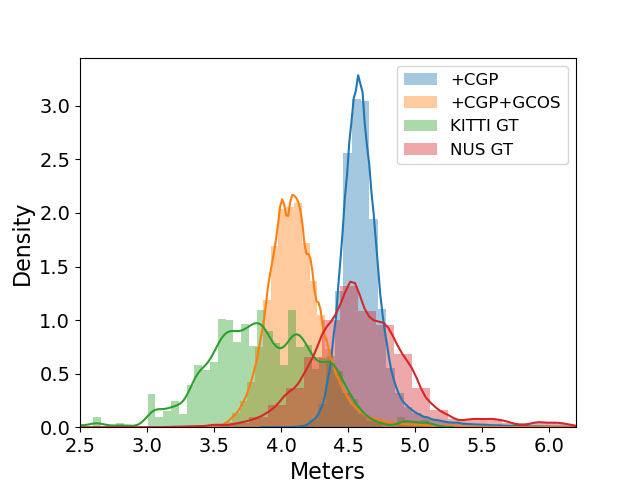}&
        \includegraphics[width=0.3\linewidth]{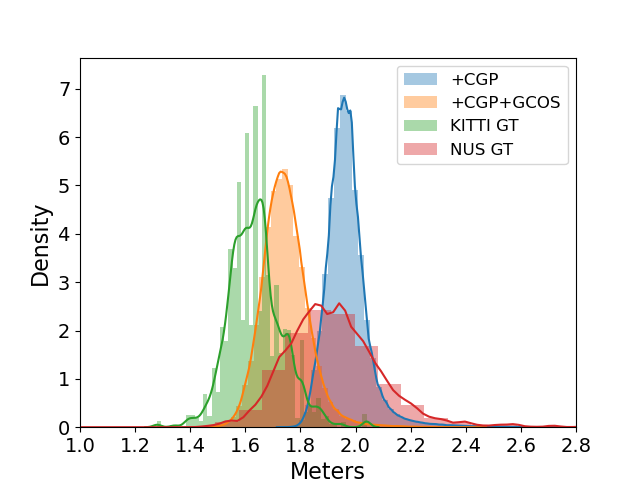}\\
        \scriptsize{(a) Height} & \scriptsize{(b) Length} & \scriptsize{(c) Width} \\
    \end{tabular}}
    \caption{Size statistics of different datasets and model predictions (Nus $\rightarrow$ KITTI).  The distribution of object sizes varies drastically across datasets. Models trained in the source domain without GCOS predict the object size with bias. Our proposed GCOS alleviates this issue, achieving more generalized Mono3D detectors.} 
    \label{fig::statistics}
\end{figure}

\subsection{Overview}
\label{sec::subsec::overview}
As shown in Fig.~\ref{fig::overall&pit}\textcolor[rgb]{1,0,0}{a}, DGMono3D is a two-stage training method based on standard center-based Mono3D detectors that predict 3D attributes including classes, 2D locations, depths, sizes, and directions. The input images are first passed through the \textit{camera-generalized paradigm} (CGP) consisting of the position-invariant transform (PIT)~\cite{gu2021pit} and the multi-scale augmentation (MS)~\cite{li2022stmono3d,park2021dd3d,chen2022graphdetr3d}. The former projects the image onto a spherical surface to eliminate the image distortion caused by FOV~\cite{gu2021pit}, and the latter combines the modification of \textit{pixel-size depth} to learn the invariant geometry correspondences in scaled images and camera parameters~\cite{li2022stmono3d}. In the first training stage, we initialize the detector encoder with standard classification pre-trained parameters and prohibit the 2D-3D \textit{geometry-consistent object scaling strategy} (GCOS) to prevent from lack of location information. After obtaining models with enough detection capability, we fine-tune the detectors with the GCOS to enhance the model generalization.

\subsection{Camera-Generalized Paradigm}
\label{sec::subsec::cgp}

As shown in Fig.~\ref{fig::overall&pit}, the vanilla CGP first randomly scales images and camera parameters as
\begin{equation}
    \mathbf{K} =
    \left[\begin{array}{ccc} r_x & r_y & 1\end{array} \right]
    \left[\begin{array}{ccc} f_x & 0 & p_x\\
                             0 & f_y & p_y\\
                             0 & 0 & 1\end{array} \right],
\end{equation}
where $r_x$ and $r_y$ are resize rates, $f$ and $p$ are focal length and optical center, $x$ and $y$ indicate image coordinate axises, respectively. Since the scaling operation cannot change FOV (proofed in the \textit{supplementary material}), the model cannot be FOV-aware, thus suffering from the influence of this discrepancy when conducting cross-dataset inference. Hence, we apply the position-invariant transform (PIT)~\cite{gu2021pit} to diminish the FOV distortion and make the model more generalized. Given the coordination $(X, Y)$ of images, we project it to the spherical coordination $(U, V)$ by
\begin{equation}
\label{eq::pit}
    U(X) = f_x\times \arctan{\left(\frac{X}{f_x}\right)},
    ~~~V(Y) = f_y\times \arctan{\left(\frac{Y}{f_y}\right)},
\end{equation}
where bilinear interpolation is adopted to sample points from images. More details can refer to~\cite{gu2021pit}. However, since the PIT is time-consuming and has to save transformed images in an offline manner, it leads to unacceptable overhead costs and makes tuning scaling parameters much harder. To solve this issue, we reverse the order of the MS and PIT based on the reversibility proofed in the \textit{supplementary material}. Hence, we can apply MS in an online manner, which saves physical memory by $n$ times when there are $n$-scale transformations as illustrated in Fig.~\ref{fig::overall&pit}\textcolor[rgb]{1,0,0}{b}.

On top of that, we replace the \textit{metric depth} $d_g$ with the \textit{pixel-size depth} $d_p$ following~\cite{li2022stmono3d,park2021dd3d}:
\begin{equation}
\label{eq::convert}
    d_p = \frac{s}{c}\cdot d_g,
    ~~s = \sqrt{\frac{1}{f_x^2} + \frac{1}{f_y^2}},
\end{equation}
where $s$ and $c$ are the pixel size and a constant, $d_p$ is the model prediction which is scaled to the final result $d_g$ with Eq.~\ref{eq::convert}. The MS strategy and the pixel-size depth make model camera-aware~\cite{li2022stmono3d,park2021dd3d}, ensuring the basic model generalization on other target domains with different camera devices.

In addition, since the pixels generalized in PIT-transformed images occupy different sizes of areas in the plain images, they have different pixel sizes which can be computed through:
\begin{equation}
\label{eq::diffsize1}
    w_x(U) = X(U+1) - X(U), ~~w_y(V) = Y(V+1) - Y(V),
\end{equation}
\begin{equation}
\label{eq::diffsize2}
    s = \sqrt{\left(\frac{w_x}{f_x}\right)^2 + \left(\frac{w_y}{f_y}\right)^2},
\end{equation}
where $(w_x, w_y)$ is real side length of PIT-transformed pixels. $s$ is vairous for pixels with different positions on images. We present the visualized weight map and give more intuitive description in the \textit{supplementary material}.


\begin{figure}[t]
    \centering
    \footnotesize
    \setlength{\tabcolsep}{10pt}
    \scalebox{0.99}{%
    \begin{tabular}{cc}
        \multicolumn{2}{c}{\includegraphics[width=0.95\linewidth]{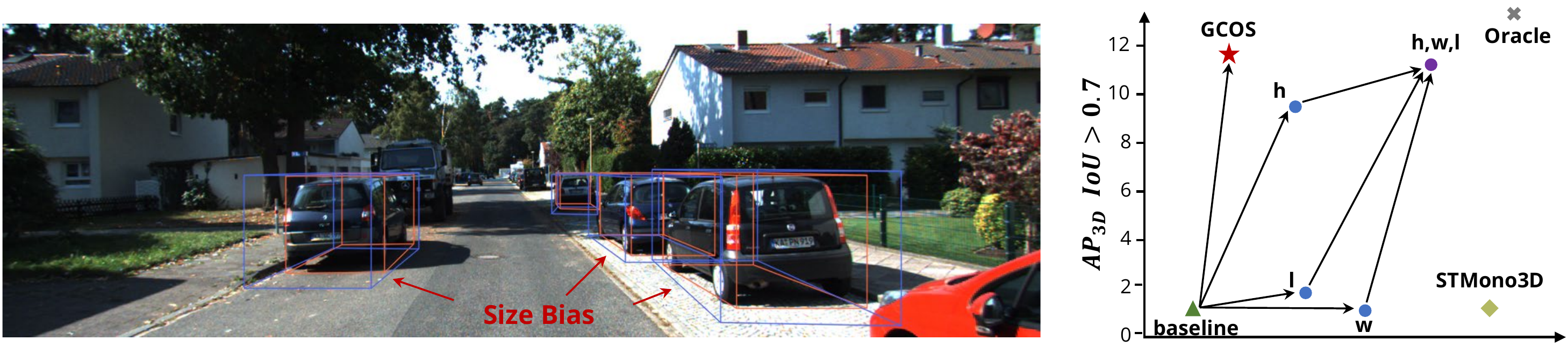}}\\
        \hspace{0.22\linewidth}\scriptsize{(a) Visualization of Size Bias} & 
        \hspace{0.2\linewidth}\scriptsize{(b) Systematic Analysis} \\
    \end{tabular}}
    \caption{(a) The pretrained model on NuScenes dataset predicts objects with larger size in the target KITTI dataset, indicating a potential size bias. The ground-truth and model prediction are presented in \textcolor[rgb]{0.23,0.4,1}{blue} and \textcolor[rgb]{0.94,0.39,0.28}{orange}, respectively. (b) We further evaluate the results by replacing the predicted dimensions of object sizes with ground-truth value step by step to better analyze the influence of the size bias on quantitative metrics.}
    \label{fig::upperbound}
\end{figure}

\subsection{2D-3D Geometry-Consistent Object Scaling Strategy}
\label{sec::subsec::gcos}

\subsubsection{Systematic Analysis}
We take NuScenes as the source dataset and KITTI as the target dataset. From Fig.~\ref{fig::statistics} and Fig.~\ref{fig::upperbound}\textcolor[rgb]{1,0,0}{a}, our key insights include two aspects. First, the distribution of object sizes varies across datasets, leading to a geometric mismatch that can be a factor for the domain gap~\cite{luo2021multialign}. Second, directly applying a model trained on NuScenes to KITTI (referred to the +GCP in the Fig.~\ref{fig::upperbound}\textcolor[rgb]{1,0,0}{b}) is ineffective since the model predicts object sizes close to the source domain. Then, we conduct the systematic analysis by replacing the predicted sizes with the ground-truth value step by step. The results shown in Fig.~\ref{fig::upperbound}\textcolor[rgb]{1,0,0}{b} demonstrate the influence of size prediction on strict quantitative metrics. Specifically, when we replace the predicted height with the ground-truth height, the AP$_{3D}$ IOU$>$0.7 is improved from 0.6\% to 9.8\%, indicating the height plays an essential role. Subsequently, we replace all the predicted dimensions (\ie, height, width, and length) and the accuracy reaches 10.3\%, which meets the results provided by Oracle models. Therefore, the low accuracy on strict metrics is mainly caused by the size error. To alleviate it, we propose the 2D-3D \textit{geometry-consistent object scaling} strategy (GCOS).

\subsubsection{GCOS Design}
Motivated by~\cite{yang2021st3d,wang2020train}, we fine-tune the already trained object detector with the GCOS so that its predicted box sizes can better match the target statistics. The principle is to maintain the 2D-3D correspondence and avoid breaking geometry consistency, which is more complex and essential for Mono3D. To facilitate the augment, we scale objects in the 3D space and then adjust corresponding areas in 2D images, keeping the bird’s eyes view (BEV) position of the visible face invariant.

\begin{figure}[t]
    \centering
    \footnotesize
    \includegraphics[width=0.98\linewidth]{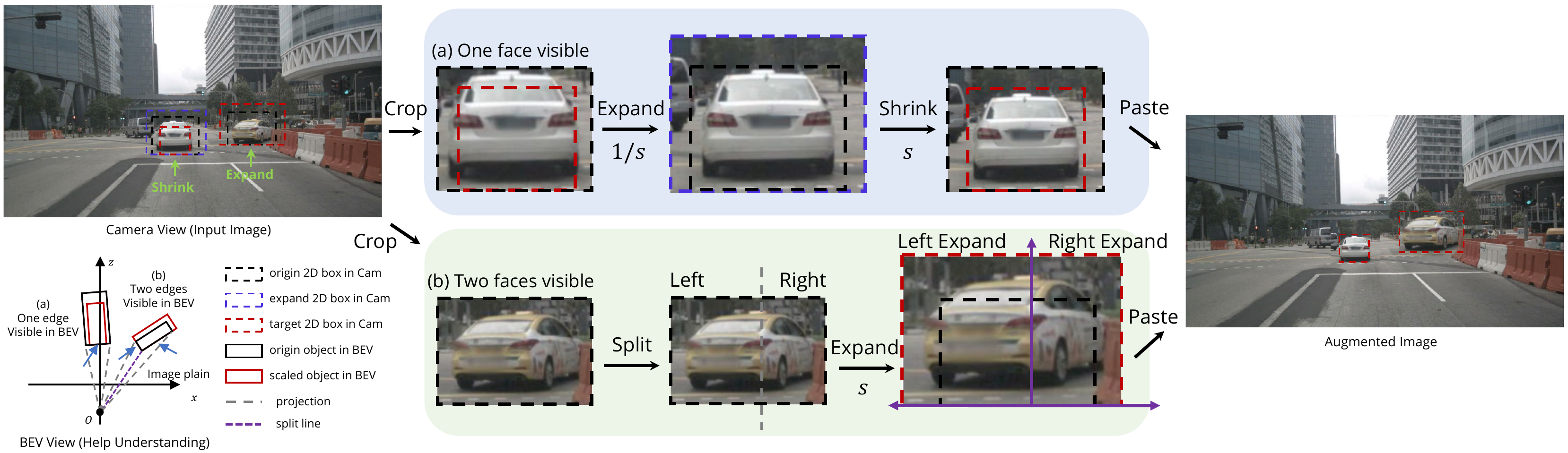}
    \caption{Illustration of the GCOS strategy. We classify objects into \textbf{(a)} objects with only \textit{one face visible} and \textbf{(b)} objects with \textit{two faces visible} (Best view in BEV). We apply two different scaling strategies on them, but the same principle is to keep the position of the BEV visible edges invariant, preventing from breaking perspective geometry consistency.} 
    \label{fig::gcos}
\end{figure}

As shown in Fig~.\ref{fig::gcos}, we classify objects into \textbf{(a)} objects with only \textit{one face visible} and \textbf{(b)} objects with \textit{two faces visible} in perspective view, and apply different strategies to scale them. As for \textbf{(a)}, we extend or shrink objects along the direction perpendicular to the visible BEV edges and retain the BEV position of visible-edge centers. In terms of \textbf{(b)}, we first split the two visible faces based on the nearest vertical fringes of 3D bounding boxes and then extend or shrink objects along the direction paralleling these two BEV visible edges. When scaling, the bottom of 3D objects is fixed on the ground. After obtaining scaled 3D bounding boxes, we project them onto 2D images, getting target boundaries. Then, we scale cropped object patches to the target size and paste them into images.

Given the dense regular arrangement of image pixels, we highlight several points in operating 2D crops. (1) On scaling objects, direct shrinking 2D crops will lead to blank fringes on origin images. To avoid leaking object information, supposing shrink objects with ratio $s$, we first expand the range of the 2D crop with $1/s$ and then narrow the expanded crop with $s$. Hence, the black fringes are filled with the background on the fly, and the objects are shrunk correctly. (2) We crop and paste objects in an inverse-depth order, avoiding breaking the layout in perspective view. (3)  To reduce artifacts caused by image patches, we follow~\cite{dwibedi2017cut,zhang2020moca} to apply random blending to smooth the boundaries of image patches. We present the installation of GCOS in Sec.~\ref{sec::subsec::implementation}.

\section{Experiments}

\begin{table}[t]
    \centering
    \caption{Dataset overview. We focus on properties related to frontal-view cameras. Sample refers to the number of images used in the training stage. We present the mean of object sizes in meters.}
    \vspace{1mm}
    \scalebox{0.8}{%
    \begin{tabular}{|c|c|c|c|c|c|c|c|}
        \hline
        ~~Dataset~~  & ~Samples~ & ~~~Loc.~~~ & ~~~~Shape~~~~ & ~~~~FOV~~~~ & ~~Height~~ & ~~Width~~ & ~~Length~~ \\
        \hline
        KITTI~\cite{geiger2012kitti} & 3712 & EUR. & (375,1242) & (29$^{\circ}$,81$^{\circ}$) & 1.52 & 1.63 & 3.87\\
        NuScenes~\cite{caesar2020nusc} & 27522 & SG.,EUR. & (900,1600) & (39$^{\circ}$,65$^{\circ}$) & 1.71 & 1.92 & 4.62\\
        Lyft~\cite{kesten2019lyft} & 21623 & SG.,EUR. & (1024,1224) & (60$^{\circ}$,70$^{\circ}$) & 1.73 & 1.94 & 4.77 \\
        \hline
    \end{tabular}
    }
\label{tab::datainfo}
\vspace{-0.2cm}
\end{table}

\subsection{Setup}
\label{sec::subsec::setup}

\textbf{Dataset.} Following~\cite{li2022stmono3d}, we conduct experiments on three widely used autonomous driving datasets: KITTI~\cite{geiger2012kitti} (CC BY-NC-SA 3.0), NuScenes ~\cite{caesar2020nusc} (CC BY-NC-SA 4.0), and Lyft~\cite{kesten2019lyft} (CC BY-NC-SA 4.0). We explore two difficulties lying in domain generalization: (1) images captured by different camera devices (\ie, pixel size and FOV), and (2) bias in the distribution of object sizes on the source and target domain. To highlight these discrepancies, we summarize the dataset information in detail at Tab.~\ref{tab::datainfo}. For Lyft and NuScenes, we subsample 1/4 data for simplicity.

\textbf{Comparison Methods.} In our experiments, we compare our DGMono3D with three counterparts: (1) \textbf{Source Only} indicates directly evaluating the source domain trained model on the target domain. (2) \textbf{Oracle} indicates the fully supervised model trained on the target domain. (3) \textbf{STMono3D}~\cite{li2022stmono3d} is the SoTA unsupervised domain adaptation method that must utilizes images on the target domain during the training stage.

\textbf{Metrics.} We adopt the KITTI evaluation metrics to evaluate our methods. Following the most domain adaptation methods for 3D detection~\cite{yang2021st3d,luo2021multialign,li2022stmono3d}, we focus on the commonly used car category. When KITTI is considered, we report the average precision (AP) where the IoU thresholds are 0.5/0.7 (\ie, loose/strict) for both the bird's eye view (BEV) IoUs and 3D IoUs. When inferring on NuScenes, since the attribute labels are different or unavailable on the target domain, we discard the average attribute error (mAAE) and report the average trans error (mATE), scale error (mASE), orient error (mAOE), and average precision (mAP). Following~\cite{yang2021st3d,li2022stmono3d}, we highlight the closed performance gap between Source Only and Oracle for a more intuitive comparison.

\subsection{Implementation}
\label{sec::subsec::implementation}

\textbf{Training.} We train DGMono3D based on FCOS3D~\cite{wang2021fcos3d} in a two-stage manner. Following~\cite{gu2021pit}, we adopt the loss re-weighting strategy to substitute the reversed PIT during the training process, which avoids the extra computational cost~\cite{gu2021pit}. In the first-stage training, we follow the standard scheme proposed in~\cite{wang2022pgd} (FCOS3D++). We train our model for 48 epoches on the KITTI dataset, and 12 epoches on the NuSenese and Lyft datasets. In the second-stage training, we fine-tune the model with the same training scheme but adopt the GCOS strategy for 2 epoches, fixing model parameters except for the size prediction branch. When the statistical information of the target domain is available, we apply GCOS with the ratio of the mean object size of the target domain to the source domain accordingly. When it is inaccessible, we utilize a random scale factor in GCOS to flatten the distribution of predicted object sizes and enhance the model generalization. Compared with the demand of collecting tremendous images on the target domain for self-teaching~\cite{li2022stmono3d}, DGMono3D is more practicable and hyper-parameter free, avoiding tricky and complex training strategies.

\begin{table}[t]
    \centering
    \caption{Performance of DGMono3D on three source-target pairs. We report AP of the car category at $\text{IoU} = 0.5/0.7$ as well as the domain gap closed by DGMono3D. In Lyft$\rightarrow$Nus, DGMono3D achieves slightly better results compared with the Oracle model on AOE, demonstrating the effectiveness of our proposed method. Moreover, our approach surpasses STMono3D~\cite{li2022stmono3d} with large margins and approach Oracle results on strict metrics, indicating the size gap is significantly bridged.}
    \vspace{1mm}
    \scalebox{0.71}{%
    \begin{tabular}{|c|ccc|ccc|ccc|ccc|}
        \hline
        \textbf{Nus$\rightarrow$K} & \multicolumn{6}{c|}{Loose Metrics} & \multicolumn{6}{c|}{Strict Metrics} \\\hline
        \multirow{2}{*}{Method}&\multicolumn{3}{c|}{$\text{AP}_{BEV}$ $\text{IoU}\geqslant0.5$} & \multicolumn{3}{c|}{$\text{AP}_{3D}$ $\text{IoU}\geqslant0.5$} & \multicolumn{3}{c|}{$\text{AP}_{BEV}$ $\text{IoU}\geqslant0.7$} & \multicolumn{3}{c|}{$\text{AP}_{3D}$ $\text{IoU}\geqslant0.7$} \\
        & Easy     & Mod.     & Hard   & Easy     & Mod.     & Hard & Easy     & Mod.     & Hard  & Easy     & Mod.     & Hard \\ 
        \hline
        Source Only & 0 & 0 & 0 & 0 & 0 & 0 & 0 & 0 & 0 & 0 & 0 & 0 \\
        Oracle & 38.77 & 30.84 & 29.82 & 34.78 & 27.97 & 26.59 & 19.98 & 16.83 & 16.29 & 15.80 & 13.61 & 13.16\\
        STMono3D~\cite{li2022stmono3d} & 35.63 & 27.37 & 23.95 & 28.65 & 21.89 & 19.55 
        & 5.37 & 4.38 & 3.76 & 0.60 & 0.64 & 0.64 \\
        DGMono3D & 34.22& 28.99& 27.82&28.77& 24.82& 23.67& 16.07& 14.89& 14.32& 12.00& 11.29& 10.97 \\
        \hline
        \cellcolor{LightCyan}Closed Gap & \cellcolor{LightCyan}88.3\% & \cellcolor{LightCyan}94.0\% & \cellcolor{LightCyan}93.2\% & \cellcolor{LightCyan}82.7\% & \cellcolor{LightCyan}88.7\% & \cellcolor{LightCyan}89.1\% & \cellcolor{LightCyan}80.4\% & \cellcolor{LightCyan}88.4\% & \cellcolor{LightCyan}87.9\% & \cellcolor{LightCyan}75.9\% & \cellcolor{LightCyan}82.9\% & \cellcolor{LightCyan}88.3\%\\
        \hline
        \hline
        \textbf{L$\rightarrow$K} & \multicolumn{3}{c|}{Loose Metrics} & \multicolumn{3}{c|}{Strict Metrics} &\multicolumn{2}{c|}{\textbf{L$\rightarrow$Nus}} & \multicolumn{4}{c|}{~~~~~~~~~~~~~~~~Metrics~~~~~~~~~~~~~~~~~~}\\
        \hline
        \multirow{2}{*}{Method}&\multicolumn{3}{c|}{$\text{AP}_{3D}$ $\text{IoU}\geqslant0.5$} & \multicolumn{3}{c|}{$\text{AP}_{3D}$ $\text{IoU}\geqslant0.7$} & \multicolumn{2}{c|}{\multirow{2}{*}{Method}} & \multirow{2}{*}{~~~AP~~~} & \multirow{2}{*}{~~ATE~~} & \multirow{2}{*}{~~ASE~~} & \multirow{2}{*}{~~AOE~~} \\
        & Easy     & Mod.     & Hard   & Easy     & Mod. & Hard & \multicolumn{2}{c|}{} &&&& \\ 
        \hline
        Source Only & 0 & 0 & 0 & 0 &  0 & 0 & \multicolumn{2}{c|}{Source Only} & 2.40 & 1.302 & 0.190 & 0.802 \\
        Oracle  & 34.78 & 27.97 & 26.59 & 15.80 & 13.61 & 13.16 & \multicolumn{2}{c|}{Oracle} & 28.2 & 0.798 & 0.160 & 0.209 \\ 
        STMono3D~\cite{li2022stmono3d} & 18.14 & 13.32 & 11.83 & 4.54 & 4.54 & 4.54 & \multicolumn{2}{c|}{STMono3D~\cite{li2022stmono3d}} & 21.3 & 0.911 & 0.170 & 0.355\\
        DGMono3D & 30.03 & 23.38 & 22.23 & 11.56 & 10.55 & 10.31 & \multicolumn{2}{c|}{DGMono3D} & 25.5 & 0.842 & 0.169 & 0.208\\
        \hline
        \cellcolor{LightCyan}Closed Gap & \cellcolor{LightCyan}86.3\% & \cellcolor{LightCyan}83.5\% & \cellcolor{LightCyan}83.6\% & \cellcolor{LightCyan}73.1\% & \cellcolor{LightCyan}77.5\% & \cellcolor{LightCyan}68.8\% & \multicolumn{2}{c|}{\cellcolor{LightCyan}Closed Gap} & \cellcolor{LightCyan}90.4\% & \cellcolor{LightCyan}91.2\% & \cellcolor{LightCyan}70.0\% & \cellcolor{LightCyan}100\%\\
        \hline
    \end{tabular}
    }
\label{tab::overall}
\vspace{-0.2cm}
\end{table}

\textbf{Inference.} When inferring on the target domain, unlike STMono3D~\cite{li2022stmono3d}, which first scales images to a certain range, we adopt images with original resolution converted by PIT~\cite{gu2021pit} for fair comparison with Oracle models. Our model outputs (\ie, 2D centers) are reversed through inv-PIT to corresponding positions on plain images~\cite{gu2021pit} and then back-project to 3D space combining the predicted object depth with camera intrinsic parameters~\cite{wang2021fcos3d,wang2022pgd}. When conducting the system analysis introduced in Sec.~\ref{sec::subsec::gcos}, we first associate the predicted and ground-truth objects via the maximum of 3D IoU, and then replace each individual object size with the ground-truth value.

\subsection{Main Results}

As shown in Tab.~\ref{tab::overall}, we compare the proposed DGMono3D with \textbf{Source Only}, \textbf{Oracle}, and the SoTA unsupervised domain adaptation method for Mono3D (\ie, \textbf{STMono3D}~\cite{li2022stmono3d}). Caused by the domain gap, the Source Only model cannot correctly localize 3D objects, and the mAP almost drops to 0\% on the target domain. In contrast, benefiting from the CGP, our DGMono3D can work on the target domain as normal, even though the images are captured by different camera devices. Specifically, DGMono3D improves the performance of Source Only models by a large margin that around 88\% and 83\% performance gaps of loose AP$_{3D}$ are closed on the NuScenes$\rightarrow$KITTI and Lyft$\rightarrow$KITTI settings. Notably, the AOE of DGMono3D even slightly better than that of Oracle on the Lyft$\rightarrow$NuScenes setting, which indicates the effectiveness of our method.

\begin{figure}[t]
    \centering
    \footnotesize
    \setlength{\tabcolsep}{10pt}
    \scalebox{0.99}{%
    \begin{tabular}{cccc}
        \multicolumn{4}{c}{\includegraphics[width=0.96\linewidth]{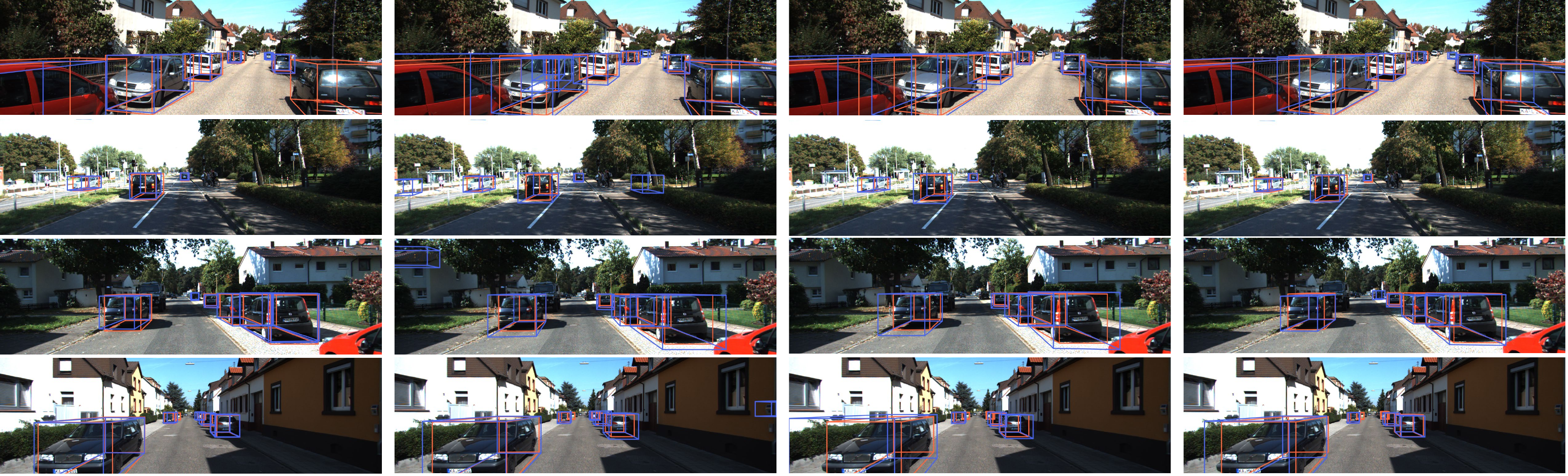}}\\
        \hspace{0.083\linewidth}\scriptsize{(a) Oracle} &  \hspace{0.083\linewidth}\scriptsize{(b) STMono3D~\cite{li2022stmono3d}} &  \hspace{0.083\linewidth}\scriptsize{(c) CGP} &  \hspace{0.07\linewidth}\scriptsize{(b) GCOS} \\
    \end{tabular}}
    \caption{We present quantitative comparisons based on the Nus$\rightarrow$KITTI setting. Benefiting from CGP, DGMono3D fully considers the FOV and pixel-size discrepancies among images captured by different camera devices, achieving more generalized models with better performance compared with STMono3D~\cite{li2022stmono3d} on the target domain. Moreover, with GCOS, the size bias is significantly alleviated, yielding more tight 3D object predictions. Zoom in for clear comparison.} 
    \label{fig::vis-compare}
\end{figure}

Moreover, DGMono3D achieves superior performance than STMono3D even without training models with data on the target domain. We highlight the strict metrics on the KITTI dataset. Although STMono3D successfully bridges a large part of the gaps on the loose metrics, its performance on strict metrics is hampered by the size gap as discussed in~\ref{sec::subsec::gcos}. In terms of the DGMono3D, the size bias is alleviated by the proposed GCOS, improving the model performance on strict metrics by significant margins. To better validate the effectiveness of DGMono3D, we present quantitative comparisons in Fig.~\ref{fig::vis-compare}. With GCOS, the size bias is significantly alleviated, achieving more tight 3D object predictions, as shown in the figures.

Although~\cite{yang2021st3d,li2022stmono3d} show that label annotations on the Lyft dataset are not comprehensive compared with the KITTI and NuScenes datasets, our DGMono3D achieves satisfying results and avoids drastic performance degradation as STMono3D~\cite{li2022stmono3d}, demonstrating the better generalization of our method. Specifically, the large discrepancy of FOV among Lyft and other datasets may lead to such a difficulty when cross-dataset inference, but our CGP can alleviate it via the embedded PIT~\cite{gu2021pit}.
 
\subsection{Ablation Studies and Discussions}

We present ablation study results to demonstrate the effectiveness of each component in DGMono3D and further discuss the reasons for the ups and downs of performance. More results and discussions about methods can be found in the \textit{supplementary material}.

\begin{table}[t]
    \centering
    \caption{Effectiveness of our proposed \textit{camera-generalized paradigm}, including the position-invariant transform (PIT)~\cite{gu2021pit} and multi-scale training with the pixel-size depth strategy (MSPSD)~\cite{li2022stmono3d}. We further investigate each component and detailed training settings in a more fine-grained manner.}
    \vspace{1mm}
    \scalebox{0.74}{%
    \begin{tabular}{|c|ccc|ccc|ccc|}
        \hline
        \textbf{Nus$\rightarrow$K} & \multicolumn{3}{c|}{Method} & \multicolumn{3}{c|}{Loose Metrics} & \multicolumn{3}{c|}{Strict Metrics} \\\hline
        \multirow{2}{*}{Domain}& \multicolumn{2}{c}{~~PIT~~} & \multirow{2}{*}{MSPSD} &  \multicolumn{3}{c|}{$\text{AP}_{3D}$ $\text{IoU}\geqslant0.5$} & 
        \multicolumn{3}{c|}{$\text{AP}_{3D}$ $\text{IoU}\geqslant0.7$} \\ 
        & w/ diff. & w/o diff. & & Easy  & Mod.  & Hard  & Easy  & Mod.  & Hard \\ 
        \hline
        \multirow{6}{*}{Source Only} & & & & 0 & 0 & 0 & 0 & 0 & 0\\
        & \ding{51} &  &   & 0.74 & 1.26 & 1.28 & 0.11 & 0.33 & 0.33\\
        &  & \ding{51} &   & 0.82 & 1.42 & 1.43 & 0.21 & 0.37 & 0.37\\
        & & & \ding{51}  & 20.75 & 18.99 & 19.24 & 0.30 & 0.26 & 0.27\\
        & \ding{51} &  & \ding{51} & 21.98 & 22.63 & 22.10 & 0.61 & 0.57 & 0.54\\
        & & \ding{51} & \ding{51}  & \cellcolor{LightCyan}27.31 & \cellcolor{LightCyan}23.08 & \cellcolor{LightCyan}22.31 & \cellcolor{LightCyan}0.73 & \cellcolor{LightCyan}0.67 & \cellcolor{LightCyan}0.66 \\
        \hline
        \multirow{5}{*}{Target (Oracle)} &  &  & & 34.78 & 27.97 & 26.59 & 15.80 & 13.61 & 13.16\\
        & \ding{51} &  & & 30.16 & 23.13 & 21.49 & 10.60 & 6.12 & 5.54 \\
        & & \ding{51} & & 29.19 & 25.87 & 24.08 & 8.75 & 5.70 & 5.01 \\
        &  &  & \ding{51} & 27.18 & 22.94 & 20.72 & 12.09 & 11.03 & 10.59 \\
        & \ding{51} &  & \ding{51} & 35.85 & 27.68 & 24.54 & 10.49 & 7.86 & 7.05 \\
        \hline
    \end{tabular}
    }
\label{tab::abl-CGP}
\vspace{-0.2cm}
\end{table}

\subsubsection{Effectiveness of the CGP}
We investigate the effectiveness of CGP, including the position-invariant transform (PIT) and the multi-scale training with the pixel-size depth strategy (MSPSD), on both target and source domains. Detailed ablation results are presented in Tab.~\ref{tab::abl-CGP}.

\textbf{PIT.} As for the oracle models, the PIT cannot improve the model performance but leads to a slight decline. We argue that the elimination of geometry cues caused by PIT hampers the performance improvement. The distortion that is removed via the PIT conversion around images can be useful information for Mono3D detectors to localize objects. Moreover, the results (\textit{w/ diff.} \vs, \textit{w/o diff.}) also indicate the application of different pixel sizes (Eq.~\ref{eq::diffsize1}, \ref{eq::diffsize2}) on PIT-transformed images is essential for geometry consistency in Mono3D.

In terms of the single-domain generalization (Source Only), the mere installation of the PIT cannot help Mono3D detectors localize objects given images captured by different camera devices on the source domain since it cannot solve the depth-issue proposed in~\cite{li2022stmono3d}. However, combining the MSPSD, CGP boosts the model performance on the target domain with a significant margin, indicating the effectiveness of our proposed CGP. One different experiment result on Source Only models is that model performance drops when we apply the different pixel sizes. One of the possible reasons is the discrepancy of image resolution as presented in Tab.~\ref{tab::datainfo}, which leads to additional domain gaps and hampers the model generalization. Therefore, we utilize the constant pixel size calculated from plain images in converting the pixel-size depth to metric depth.

\begin{table}[t]
    \centering
    \caption{Effectiveness of our proposed geometry-consistent object scaling strategy.}
    \vspace{1mm}
    \scalebox{0.78}{%
    \begin{tabular}{|c|c|ccc|ccc|c|c|c|}
        \hline
         & &  \multicolumn{6}{c|}{\textbf{Nus$\rightarrow$K}} &   \multicolumn{2}{c|}{\textbf{Nus$\rightarrow$Lyft}} \\\hline
        \multirow{2}{*}{Stat. Info.} & \multirow{2}{*}{Method} & 
        \multicolumn{3}{c|}{$\text{AP}_{BEV}$ $\text{IoU}\geqslant0.7$} & 
        \multicolumn{3}{c|}{$\text{AP}_{3D}$ $\text{IoU}\geqslant0.7$} & 
        $\text{AP}_{BEV}$  &
        $\text{AP}_{3D}$  \\ 
        & &  Easy  & Mod.  & Hard  & Easy  & Mod.  & Hard & $\text{IoU}\geqslant0.7$ & $\text{IoU}\geqslant0.7$ \\ 
        \hline
        &baseline & 6.94 & 6.25 & 6.11 & 0.72 & 0.66 & 0.66 & 8.83 & 4.36\\
        \ding{51}&scale pred & 15.85 & 14.71 & 14.50 & 0.99 & 0.94 & 0.92 & 8.55 & 4.19\\
        \ding{51}&gt rep.  & 16.36 & 15.13 & 14.95 & 10.71 & 10.32 & 10.21 & 9.52 & 5.11 \\
        \cellcolor{LightCyan}\ding{51}&\cellcolor{LightCyan}stat. gcos & \cellcolor{LightCyan}16.07 & \cellcolor{LightCyan}14.89 & \cellcolor{LightCyan}14.32 & \cellcolor{LightCyan}12.00 & \cellcolor{LightCyan}11.29 & \cellcolor{LightCyan}10.97 & \cellcolor{LightCyan}16.90 & \cellcolor{LightCyan}12.32 \\
        \hline
        \ding{55}&rand. gcos  & 11.30 & 10.70 & 10.48 & 1.07 & 1.02 & 1.02 & 12.07 & 7.82 \\
        \hline
        &Oracle  & 19.98 & 16.83 & 16.29 & 15.80 & 13.61 & 13.16 & 19.19 & 8.52\\
        \hline
    \end{tabular}
    }
\label{tab::abl-GCOS}
\vspace{-0.2cm}
\end{table}

\textbf{MSPSD.}
MSPSD plays a crucial role in domain generalization. As discussed in~\cite{li2022stmono3d}, it forces the Mono3D detectors to predict object depth via cues of object acreages in perspective view. With the help of MSPSD, model generalization is improved significantly. However, during the training stage, other depth cues are inevitably weakened, leading to a slight decrease in model performance for Oracle models. 


\subsubsection{Effectiveness of the GCOS}
\label{sec::subsec::subsubsec::abl-gcos}

We choose the trained model equipped with DGP as the baseline for the ablation study of GCOS. We compare the following strategies to investigate the effectiveness of GCOS. Supposing the statistical ratio of the target domain to the source domain is $s_t$. (1) \textit{scale pred.}: directly multiply size predictions with $s_t$ when inference on the target domain. (2) \textit{gt rep.}: replace the predicted sizes with the ground-truth value.  (3) \textit{stat. gcos}: apply GCOS with $s_t$. (4) \textit{rand. gcos}: apply GCOS with a random scaling factor $s_r$ in the condition that the statistical information is \textbf{inaccessible}. Moreover, we also consider domain generalization in two different situations with large (\eg, Nus$\rightarrow$K) and small (\eg, Nus$\rightarrow$Lyft) domain gaps of size distributions, respectively.

As shown in Tab.~\ref{tab::abl-GCOS}, no matter how large the discrepancy of size distribution exists between the source and target domain, directly applying \textit{scale pred.} \textbf{cannot} improve the model performance on strict metrics. However, as for the models fine-tuned by \textit{stat. gcos}, the degradation of performance on strict metrics is remarkably alleviated, which definitely proves the effectiveness of our GCOS. 

Furthermore, we also explore the condition that we have no statistical information on the target domain, which is a much more challenging setting. We adopt \textit{rand. gcos} to flatten the prediction distribution of size and improve the model generalization. In the tough setting of Nus$\rightarrow$K, while the strict metric of AP$_{BEV}$, IOU$\geqslant$0.7 is enhanced with a great margin, indicating GCOS with random scaling factor is beneficial to model generalization, the AP$_{3D}$, IOU$\geqslant$0.7 is on par with the baseline model, which suggests a difficulty in Mono3D detectors to predict correct object height when cross-dataset inference. However, as for the situation where the size gap is not oversized, which is a more common setting for real-world applications, \textit{rand. gcos} successfully bridge size gaps with prominent margins, indicating that GCOS can make Mono3D detectors more generalized.

\section{Conclusion}

This paper presents DGMono3D, a meticulously designed single-domain generalization framework tailored for the monocular 3D object detection task. We first combine the PIT~\cite{gu2021pit} and multi-scale training with the pixel-size depth strategy~\cite{li2022stmono3d,park2021dd3d,chen2022graphdetr3d} to construct a unified \textit{camera-generalized paradigm} that fully consider the discrepancies of FOV and pixel size when cross-dataset inference on images captured by different camera devices. Then, we investigate model performance degradation on strict quantitative metrics caused by different distributions of object sizes on the source and target domains. To alleviate this issue, we propose the 2D-3D \textit{geometry-consistent object scaling} strategy to scale objects based on statistical information during the training stage. Extensive experimental results on three datasets demonstrate the effectiveness of DGMono3D, which can serve as a solid baseline for industrial applications and further research on domain adaptation for Mono3D.

{\small
\bibliographystyle{ieee_fullname}
\bibliography{ref}
}

\appendix


\section{Proofs}

\subsection{Invariance of FOV in Image Scaling}
\label{subsec::proof_fov_invariance}
Consider the formulation of FOV:

\begin{equation}
\begin{aligned}
\nonumber
\textrm{FOV}_w &= 2 \times \arctan{\left(\frac{w}{2f_x}\right)}, \\
\textrm{FOV}_h &= 2 \times \arctan{\left(\frac{h}{2f_y}\right)},
\end{aligned}
\end{equation}
where $w$, $h$ are image width and height, $f_x$, $f_y$ are focal lengths along the $x$ (horizontal) and $y$ (perpendicular) direction, respectively.

When scaling images, we simultaneously operate the image resolution and the camera intrinsic parameter as:
\begin{equation}
\nonumber
    w' = r_x \times w, ~~~ h' = r_y \times h,
\end{equation}
\begin{equation}
\nonumber
    \mathbf{K'} =
    \left[\begin{array}{ccc} f_x' & 0 & p_x'\\
                             0 & f_y' & p_y'\\
                             0 & 0 & 1\end{array} \right] = 
    \left[\begin{array}{ccc} r_x & r_y & 1\end{array} \right]
    \left[\begin{array}{ccc} f_x & 0 & p_x\\
                             0 & f_y & p_y\\
                             0 & 0 & 1\end{array} \right],
\end{equation}
where $r_x$, $r_y$ are scaling factor and $K$ is the camera intrinsic parameter.

After scaling the image with $r_x$ and $r_y$, we re-calculate the FOV:
\begin{equation}
\begin{aligned}
\nonumber
\textrm{FOV}_w' &= 2 \times \arctan{\left(\frac{w'}{2f_x'}\right)} \\
&= 2 \times \arctan{\left(\frac{r_x \times w}{2f_x \times r_x}\right)} \\
&= 2 \times \arctan{\left(\frac{w}{2f_x}\right)} = \textrm{FOV}_w.
\end{aligned}
\end{equation}
Similarly, the invariance of FOV$_h$ can be proofed. We add more discussions about consequences caused by this invariance during multi-scale training (MS) in Sec.~\ref{subsubsec::pit}.

\subsection{Reversibility of MS and PIT}
\subsubsection{Vanilla CGP}
We present the sequence that we first apply MS and then PIT. The anchor pixel on the coordinate of $(x, y)$ changes to $(x', y')$ after scaling the image with scale factors $r_x$ and $r_y$ as:
\begin{equation}
\label{eq::x1=rxx}
    x_1 = r_x \times x,~~~y_1 = r_y \times y.
\end{equation}

The camera intrinsic parameter is changed simultaneously as:
\begin{equation}
\label{eq::k'mspit}
    \mathbf{K'} =
    \left[\begin{array}{ccc} f_x' & 0 & p_x'\\
                             0 & f_y' & p_y'\\
                             0 & 0 & 1\end{array} \right] = 
    \left[\begin{array}{ccc} r_x & r_y & 1\end{array} \right]
    \left[\begin{array}{ccc} f_x & 0 & p_x\\
                             0 & f_y & p_y\\
                             0 & 0 & 1\end{array} \right].
\end{equation}

Finally, we apply PIT to convert the pixel onto the spherical coordinate as
\begin{equation}
\label{eq::u1=f(x1,fx')}
    u_1 = f_x'\times \arctan{\left(\frac{x_1}{f_x'}\right)},
    ~~~v_1 = f_y'\times \arctan{\left(\frac{y_1}{f_y'}\right)}.
\end{equation}

\subsubsection{Advanced CGP}
In this advanced version, we first apply PIT and then MS. The anchor pixel on the coordinate of $(x, y)$ is converted onto the spherical coordinate as
\begin{equation}
\label{eq::u=f(x,fx)}
    u = f_x\times \arctan{\left(\frac{x}{f_x}\right)},
    ~~~v = f_y\times \arctan{\left(\frac{y}{f_y}\right)}.
\end{equation}

Then, we apply the MS on the scaled image and also change the camera intrinsic parameter as:
\begin{equation}
\label{eq::u2=rxu}
    u_2 = r_x \times u,
    ~~~v_2 = r_y \times v,
\end{equation}
\begin{equation}
\nonumber
    \mathbf{K'} =
    \left[\begin{array}{ccc} f_x' & 0 & p_x'\\
                             0 & f_y' & p_y'\\
                             0 & 0 & 1\end{array} \right] = 
    \left[\begin{array}{ccc} r_x & r_y & 1\end{array} \right]
    \left[\begin{array}{ccc} f_x & 0 & p_x\\
                             0 & f_y & p_y\\
                             0 & 0 & 1\end{array} \right].
\end{equation}

After that, we can calculate the corresponding position $(x_2, y_2)$ on the raw scaled plain image as
\begin{equation}
\label{eq::x2=f(u2fx')}
x_2 = f_x' \times \tan{\left(\frac{u_2}{f_x'} \right)},~~~y_2 = f_x' \times \tan{\left(\frac{v_2}{f_y'} \right)}.
\end{equation}

\subsubsection{Invariance}
We proof $u_2 = u_1$ , $v_2 = v_1$, $x_2 = x_1$, and $y_2 = y_1$ to demonstrate the invariance of MS and PIT. Since the symmetry, we present the proof of $u_2 = u_1$ and $x_2 = x_1$. The other two equations can be proofed similarly.

\textit{The proof of} \boldmath $u_2 = u_1$\unboldmath:

\begin{equation}
\begin{aligned}
\label{eq::u2=u1}
u_2 &\xlongequal[]{Eq.~\ref{eq::u2=rxu}} r_x \times u \\
&\xlongequal[]{Eq.~\ref{eq::u=f(x,fx)}} r_x \times f_x \times \arctan{\left(\frac{x}{f_x}\right)} \\
&\xlongequal[]{} r_x \times f_x \times \arctan{\left(\frac{r_x \times x}{r_x \times f_x}\right)} \\
&\xlongequal[]{Eq.~\ref{eq::x1=rxx},\ref{eq::k'mspit}} f_x' \times \arctan{\left(\frac{x_1}{f_x'}\right)} \\
&\xlongequal[]{Eq.~\ref{eq::u1=f(x1,fx')}} u_1. \\
\end{aligned}
\end{equation}

Hence, the position of the anchor point $(x, y)$ on the scaled PIT-converted image is the same in the order and inversed process.

\textit{The proof of} \boldmath $x_2 = x_1$ \unboldmath:
\begin{equation}
\begin{aligned}
\label{eq::x2=x1}
x_2 &\xlongequal[]{Eq.~\ref{eq::x2=f(u2fx')}} f_x' \times \tan{\left(\frac{u_2}{f_x'} \right)} \\
&\xlongequal[]{Eq.~\ref{eq::u2=u1}} f_x' \times \tan{\left(\frac{u_1}{f_x'} \right)} \\
&\xlongequal[]{Eq.~\ref{eq::u1=f(x1,fx')}} f_x' \times \tan{\left(\frac{f_x'\times \ \arctan{\left(\frac{x_1}{f_x'}\right)}}{f_x'} \right)} \\
&\xlongequal[]{} x_1. \\
\end{aligned}
\end{equation}

Hence, the corresponding position of the anchor pixel on the scaled PIT-converted image $(u, v)$ is the same in the order and inversed process.

Thus, the \textit{reversibility} of MS and PIT is proofed.

\begin{table}[t]
    \centering
    \caption{Performance of DGMono3D on three source-target pairs. We present more fine-grained ablation results to demonstrate the effectiveness of each component proposed in this paper.}
    \vspace{1mm}
    \scalebox{0.71}{%
    \begin{tabular}{|c|ccc|ccc|ccc|ccc|}
        \hline
        \textbf{Nus$\rightarrow$K} & \multicolumn{6}{c|}{Loose Metrics} & \multicolumn{6}{c|}{Strict Metrics} \\\hline
        \multirow{2}{*}{Method}&\multicolumn{3}{c|}{$\text{AP}_{BEV}$ $\text{IoU}\geqslant0.5$} & \multicolumn{3}{c|}{$\text{AP}_{3D}$ $\text{IoU}\geqslant0.5$} & \multicolumn{3}{c|}{$\text{AP}_{BEV}$ $\text{IoU}\geqslant0.7$} & \multicolumn{3}{c|}{$\text{AP}_{3D}$ $\text{IoU}\geqslant0.7$} \\
        & Easy     & Mod.     & Hard   & Easy     & Mod.     & Hard & Easy     & Mod.     & Hard  & Easy     & Mod.     & Hard \\ 
        \hline
        Source Only (SO) & 0 & 0 & 0 & 0 & 0 & 0 & 0 & 0 & 0 & 0 & 0 & 0 \\
        Oracle & 38.77 & 30.84 & 29.82 & 34.78 & 27.97 & 26.59 & 19.98 & 16.83 & 16.29 & 15.80 & 13.61 & 13.16\\
        SO + PIT & 34.65& 29.44& 28.90 & 23.08 & 24.82& 22.31 &
        6.94 & 6.25 & 6.11 & 0.72 & 0.66 & 0.66 \\
        SO + GCOS (w/o stat.) & 33.35 & 27.99 & 27.28 & 25.11 &  21.52 & 20.53 & 
        11.30 & 10.70 & 10.48 & 1.07 & 1.02 & 1.02 \\
        SO + GCOS (w/ stat.) & 34.22& 28.99& 27.82&28.77& 24.82& 23.67& 
        16.07& 14.89& 14.32& 12.00& 11.29& 10.97 \\
        \hline
        \hline

        \textbf{L$\rightarrow$K} & \multicolumn{6}{c|}{Loose Metrics} & \multicolumn{6}{c|}{Strict Metrics} \\\hline
        \multirow{2}{*}{Method}&\multicolumn{3}{c|}{$\text{AP}_{BEV}$ $\text{IoU}\geqslant0.5$} & \multicolumn{3}{c|}{$\text{AP}_{3D}$ $\text{IoU}\geqslant0.5$} & \multicolumn{3}{c|}{$\text{AP}_{BEV}$ $\text{IoU}\geqslant0.7$} & \multicolumn{3}{c|}{$\text{AP}_{3D}$ $\text{IoU}\geqslant0.7$} \\
        & Easy     & Mod.     & Hard   & Easy     & Mod.     & Hard & Easy     & Mod.     & Hard  & Easy    & Mod.     & Hard \\ 
        \hline
        Source Only (SO) & 0 & 0 & 0 & 0 & 0 & 0 & 0 & 0 & 0 & 0 & 0 & 0 \\
        Oracle & 38.77 & 30.84 & 29.82 & 34.78 & 27.97 & 26.59 & 19.98 & 16.83 & 16.29 & 15.80 & 13.61 & 13.16\\
        SO + CGP & 30.36 & 26.29 & 25.48 &22.97& 18.62 & 17.88 & 
        3.55 & 3.26 & 3.19 & 0.29 & 0.36 & 0.34 \\
        SO + GCOS (w/o stat.) & 36.94 & 23.40 & 22.08 & 28.00 & 15.44 & 14.30 & 6.62 & 3.60 & 3.37 & 0.92 & 0.51 & 0.51 \\
        SO + GCOS (w/ stat.) & 36.18 & 28.30 & 27.16 & 30.03 & 23.38 & 22.23 & 
        17.29 & 14.76 & 14.30 & 11.56 & 10.55 & 10.31 \\
        \hline
    \end{tabular}
    }
    
    \vspace{1mm}
    \scalebox{0.71}{%
    \begin{tabular}{|c|cccc|cccc|}
        \hline
        \multicolumn{5}{|c|}{\textbf{K$\rightarrow$Nus}} &   \multicolumn{4}{c|}{\textbf{K$\rightarrow$Lyft}} \\\hline
        \multirow{2}{*}{Method} & 
        \multirow{2}{*}{~~~AP~~~} & \multirow{2}{*}{~~ATE~~} & \multirow{2}{*}{~~ASE~~} & \multirow{2}{*}{~~AOE~~} & $\text{AP}_{BEV}$  & $\text{AP}_{3D}$ & $\text{AP}_{BEV}$  & $\text{AP}_{3D}$  \\ 
        & &  & & & $\text{IoU}\geqslant0.5$ & $\text{IoU}\geqslant0.5$ &  $\text{IoU}\geqslant0.7$ & $\text{IoU}\geqslant0.7$ \\
        \hline
        Source Only (SO) & 2.4 & 1.302 & 0.190 & 0.802 & 0 & 0 & 0 & 0\\
        Oracle & 28.2 & 0.798 & 0.160 & 0.209 & 33.00 & 29.74 & 19.19 & 8.52 \\
        SO + CGP &  18.0  & 0.847 & 0.297 & 0.441 & 6.50 & 5.37 & 0.82 & 0.08 \\
        SO + GCOS (w/o stat.) &  18.4 &  0.842 &  0.288  & 0.446 & 10.18 & 7.00  & 1.67 & 0.82 \\
        SO + GCOS (w/ stat.) &  18.2 & 0.813 & 0.184 & 0.436 & 13.02 & 9.81 & 3.68 & 1.28 \\
        \hline
        \hline
        \multicolumn{5}{|c|}{\textbf{Lyft$\rightarrow$Nus}} &   \multicolumn{4}{c|}{\textbf{Nus$\rightarrow$Lyft}} \\\hline
        \multirow{2}{*}{Method} & 
        \multirow{2}{*}{~~~AP~~~} & \multirow{2}{*}{~~ATE~~} & \multirow{2}{*}{~~ASE~~} & \multirow{2}{*}{~~AOE~~} & $\text{AP}_{BEV}$  & $\text{AP}_{3D}$ & $\text{AP}_{BEV}$  & $\text{AP}_{3D}$ \\ 
        & &  & & & $\text{IoU}\geqslant0.5$ & $\text{IoU}\geqslant0.5$  &  $\text{IoU}\geqslant0.7$ & $\text{IoU}\geqslant0.7$ \\
        \hline
        Source Only (SO) & 7.1 & 1.182 & 0.185 & 0.447 & 0 & 0 & 0 & 0 \\
        Oracle & 28.2 & 0.798 & 0.160 & 0.209 & 33.00 & 29.74 & 19.19 & 8.52 \\
        SO + CGP & 25.5 & 0.842 & 0.169 & 0.208 & 26.57 & 22.34 & 8.83 & 4.36 \\
        SO + GCOS (w/o stat.) & 25.3 & 0.844 & 0.169 & 0.213 & 31.32 & 27.58 & 12.07 & 7.82\\
        SO + GCOS (w/ stat.) & 25.4 & 0.841 & 0.166 & 0.209 & 31.68 & 28.11 & 16.90 & 12.32\\
        \hline
    \end{tabular}
    }
\label{tab::sup-overall}
\vspace{-0.2cm}
\end{table}

\section{Experimental Results}

We present all the experimental results on three-pair datasets as shown in Tab.~\ref{tab::sup-overall} and highlight several interesting observations:

(1) Based on statistical information (Specific values are presented in Tab.1 in the paper), the distributions of object dimensions on NuScenes~\cite{caesar2020nusc} and Lyft~\cite{kesten2019lyft} are similar. In other words, the mean values are approximate. However, objects on the KITTI dataset~\cite{geiger2012kitti} are much smaller, leading to a tricky and huge domain gap. Therefore, while applying GCOS (w/o stat.) on the difficult settings (\eg, Nus$\rightarrow$K) can obtain performance gain, there are still severe degradations on the strict metrics. When the dimension discrepancy is not so huge (\eg, Nus$\rightarrow$Lyft),  GCOS (w/o stat.) can achieve satisfactory results without any information on the target domain. As for GCOS (w/ stat.), it can work effectively on all these settings, bridging large gaps caused by the dimension discrepancy.

(2) Different from metrics on KITTI and Lyft datasets, the AP on NuScenes is less affected by the accuracy of dimension predictions. In contrast, the average size error (ASE) specifically measures the accuracy of dimension predictions. Hence, the GCOS mainly contributes to the ASE instead of AP when cross-dataset inference on NuScenes.

\begin{figure}[t]
    \centering
    \footnotesize
    \includegraphics[width=0.98\linewidth]{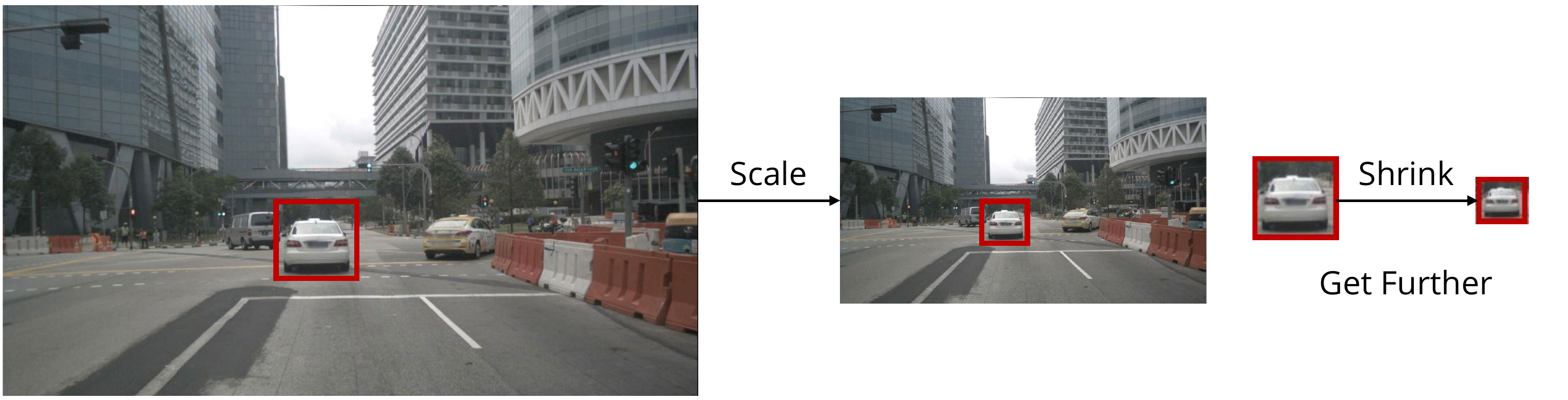}
    \caption{Illustration of the multi-scale training and pixel-size depth.} 
    \label{fig::sup-ill-ms}
\end{figure}

\section{Discussions about CGP and GCOS}

\subsection{Camera-Generalized Paradigm}
\subsubsection{What Do Models Learn from Multi-Scale Training and Pixel-Size Depth?}
\label{subsubsec::ms}

As investigated in~\cite{li2022stmono3d} and our paper, the multi-scaling training (MS) and pixel-size depth plays crucial roles in cross-domain inference. In~\cite{park2021dd3d}, the authors claim that they can make models \textit{camera-aware}, but there is still a lack of explanation in literature. Here, we aim to dig into what models learn from the MS and pixel-size depth.

As introduced in the paper, we replace the \textit{metric depth} $d_g$ with \textit{pixel-size depth} $d_p$ following~\cite{li2022stmono3d,park2021dd3d}:
\begin{equation}
\label{eq::pixel-depth}
    d_p = \frac{s}{c}\cdot d_g,
    ~~s = \sqrt{\frac{1}{f_x^2} + \frac{1}{f_y^2}}.
\end{equation}
In practice, we do not change the $d_g$ on scaled images, maintaining the invariance of the 3D space. When scaling the images and the camera intrinsic parameters, we imitate taking pictures with different cameras (\ie, intrinsic parameters). The models predict $d_p$, which is post-processed via multiplying $c/s$ to get the metric-depth prediction. 

However, it is hard to understand the function of MS since the entanglement of factor $c/s$ and pixel-size depth $d_p$. To better understand it, we can equivalently analyze the scaling of $d_g$ with factor $s/c$, whose result is the depth we want the models to predict from the scaled image (\ie, target depth). Reasonably, for an example as illustrated in Fig.~\ref{fig::sup-ill-ms}, when we shrink a image with a scale factor $r < 1$, the areas of objects are reduced. In perspective view, the scaled objects are correspondingly getting further from the camera. Since there is a negative correlation between the factor $s/c$ and the scale factor $r$, when shrinking the images with a smaller scale factor $r$, the metric depth $d_g$ will multiply with a more significant factor $s/c$ to get a more distant target depth. Therefore, we essentially enhance the depth cues that \textit{more distant objects are smaller} in perspective view via the multi-scale training. Moreover, it also increases the diversity of factor $s/c$, making the post-processing more generalized.

Since the geometry structure of 3D space is invariant among datasets, the depth cues in perspective view (\textit{more distant objects are smaller}) learned from the source domain can work as usual on the target domain, thus obtaining reasonable and correct predictions of $d_p$. Combined with the factor $c/s$ that can also be calculated from the camera intrinsic parameter, Mono3D detectors can localize object depth $d_g$ without the depth shift illustrated in~\cite{li2022stmono3d} on the target domain. If we directly adopt $d_p$, the models still predict object depth based on similar depth cues, thus leading to depth shift since the models have no ideas about the influence of different cameras on object size in perspective view. In other words, benefiting from the dataset-invariant pixel-size depth $d_p$ and the consideration of the influence of camera parameters (\ie, $c/s$) on object size, Mono3D detectors can predict correct metric-depth $d_g$ on the target domain.

\begin{figure}[t]
    \centering
    \footnotesize
    \setlength{\tabcolsep}{10pt}
    \scalebox{0.99}{%
    \begin{tabular}{cc}
        \multicolumn{2}{c}{\includegraphics[width=0.96\linewidth]{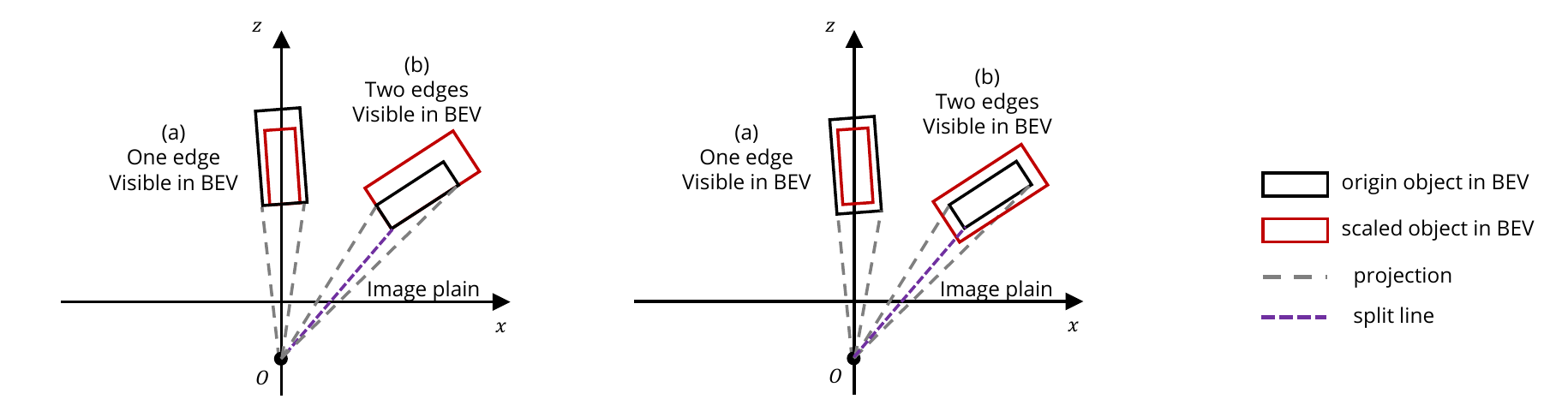}}\\
        \hspace{0.02\linewidth}\scriptsize{(a) 2D-3D geometry-consistent object scaling} & 
        \hspace{-0.3\linewidth}\scriptsize{(b) Vanilla object scaling} \\
    \end{tabular}}
    \caption{Comparison of the 2D-3D geometry-consistent object scaling and vanilla object scaling.}
    \label{fig::sup-compare-gcos}
\end{figure}

\subsubsection{Why Do We Need to Apply the Position Invariant Transform?}
\label{subsubsec::pit}

As discussed in~\ref{subsubsec::ms}, multi-scale training is trying to imitate taking pictures with different cameras (\ie, intrinsic parameters). Therefore, equipped with pixel-size depth, the model will be \textit{aware} of the discrepancy of camera parameters among datasets when cross-domain inference. However, since the multi-scale augment cannot change the camera FOV as proofed in~\ref{subsec::proof_fov_invariance}, the FOV gaps among different datasets can lead to the performance degradation. Moreover, it is tough to generate images with different FOVs for training and model the influence of FOV discrepancy on Mono3D. To alleviate the potential domain gap raised by FOV, we adopt the position invariant transform (PIT)~\cite{gu2021pit} to remove the distortion caused by FOV. The models achieve 3D detection based on images without FOV distortion, avoiding the potential domain gap raised by FOV.

\begin{figure}[t]
    \centering
    \footnotesize
    \setlength{\tabcolsep}{10pt}
    \scalebox{0.99}{%
    \begin{tabular}{cc}
        \multicolumn{2}{c}{\includegraphics[width=0.9\linewidth]{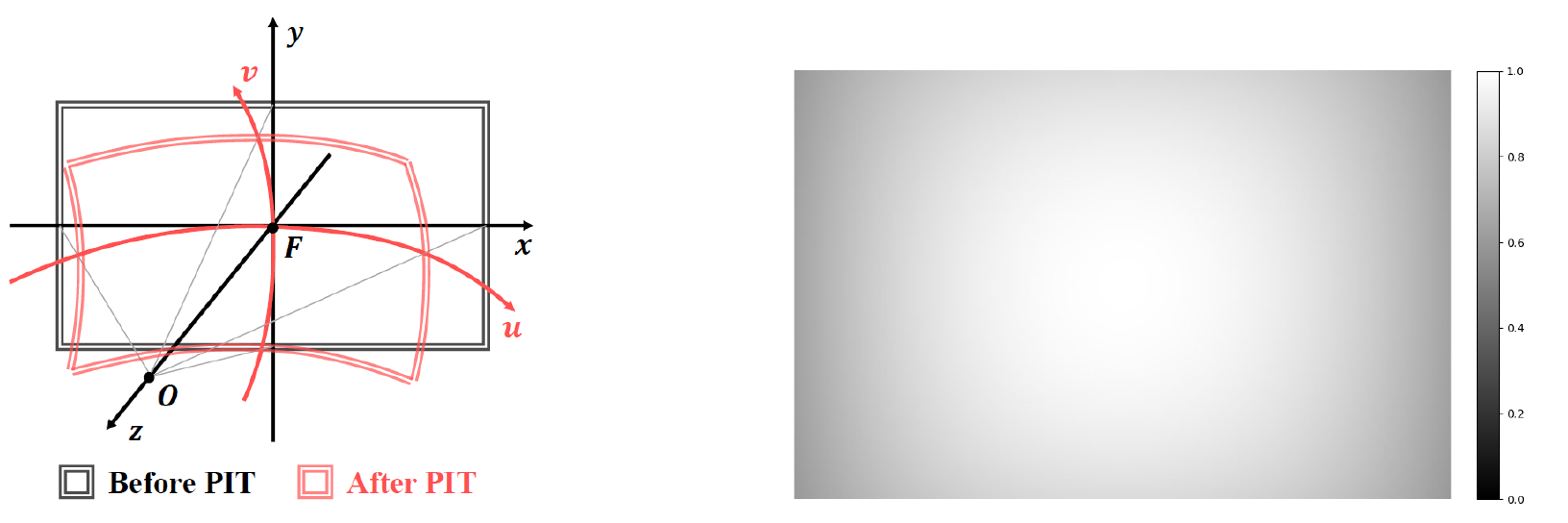}}\\
        \hspace{0.1\linewidth}\scriptsize{(a) PIT~\cite{gu2021pit}} & 
        \hspace{0.2\linewidth}\scriptsize{(b) Different scale factors $\frac{c}{s}$} \\
    \end{tabular}}
    \caption{Illustration of different scale factors for pixel-size depth.}
    \label{fig::sup-diff}
\end{figure}

\subsubsection{Visualization of Different Scale Factors}
In the paper, we have discussed the different depth scale factors in PIT-converted images: 
\begin{equation}
\begin{aligned}
\nonumber
w_x(U) = X(&U+1) - X(U), ~~w_y(V) = Y(V+1) - Y(V),\\
d_p &= \frac{s}{c}\cdot d_g,
~~s = \sqrt{\frac{w_x}{f_x^2} + \frac{w_y}{f_y^2}},
\end{aligned}
\end{equation}
Here, we present more intuitive introductions. As shown in Fig.~\ref{fig::sup-diff}\textcolor[rgb]{1,0,0}{a} from~\cite{gu2021pit}, pixels on the converted sphere coordinate occupy different size of area on the plain raw images. Hence, the pixel size is various in the PIT-converted images. To keep the geometry consistency, we apply various scale factor $\frac{c}{s}$ (shown in  Fig.~\ref{fig::sup-diff}\textcolor[rgb]{1,0,0}{b}) when converting the predicted pixel depth $d_p$ to the metric depth $d_g$.

More intuitively, at the edges of the PIT-converted image, pixels are `larger' with more corresponding pixels on the plain raw image. Considering the explanation in~\ref{subsubsec::ms}, objects in these areas look smaller compared to the ones in the center of the PIT-converted image. In perspective view, smaller means further, and the model will predict a larger depth value. Remember that we do not scale the ground-truth value during the training stage. Therefore, it makes sense that we need to multiply a smaller scale factor $\frac{c}{s}$ on the larger predicted depth to make the predicted metric depth closer to the ground truth.

\subsection{2D-3D Geometry-Consistent Object Scaling}

\subsubsection{Why is the 2D-3D Geometry Consistency?}
In LIDAR-based methods~\cite{wang2020train,yang2021st3d}, the object scaling strategy is center-based as shown in Fig.~\ref{fig::sup-compare-gcos}\textcolor[rgb]{1,0,0}{b}, which means the BEV position of the object center is unchanged. It is undoubtedly reasonable when considering that points are of discrete distribution, and the points reflected by objects are also centered on 3D centers. However, as for Mono3D, we aim to localize objects given monocular images. Directly applying the vanilla object scaling strategy will lead to the ambiguity that the object size in the 2D perspective view is changed by the combined influence of 3D-object scaling and depth variation (best think in BEV). In contrast, in our 2D-3D geometry-consistent object scaling strategy, we maintain the BEV position of the visible edges. Hence, the object size in the 2D perspective view is only changed by the influence of 3D-object scaling. While the experimental results are on par, it makes sense that we need to decrease potential factors leading to performance degradation.

\section{Limitation}
\begin{figure}[t]
    \centering
    \footnotesize
    \includegraphics[width=0.98\linewidth]{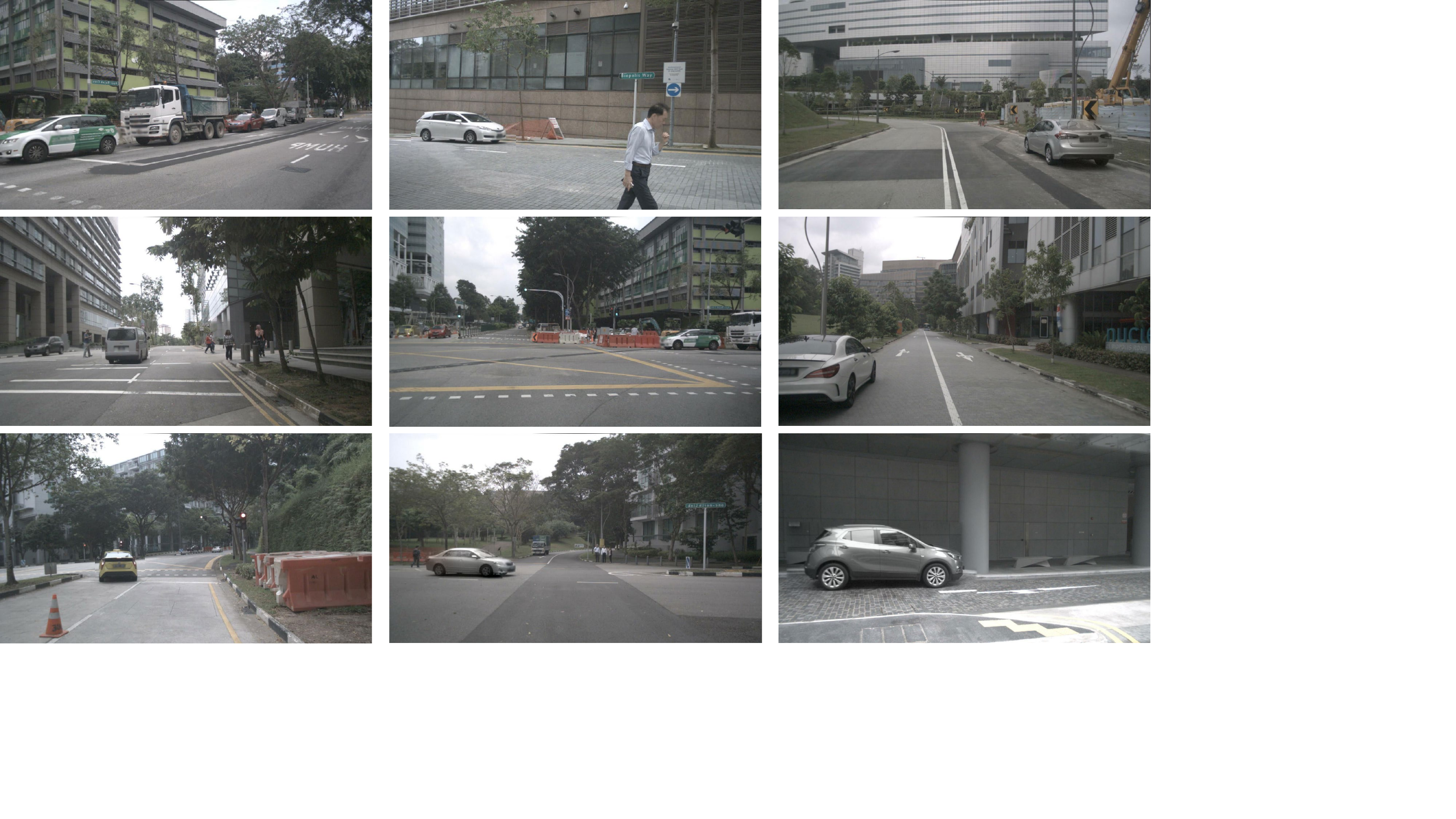}
    \caption{Images augmented by the 2D-3D GCOS.} 
    \label{fig::sup-images}
\end{figure}

We present more images augmented via 2D-3D GCOS in Fig.~\ref{fig::sup-images}. However, caused by the dense regular arrangement of image pixels, it is inevitable to introduce artifacts and leak localization information when applying the scaling strategy. We have to utilize a two-stage fine-tuning strategy to train the dimension branch of generalized models, avoiding hurting the model performance on the target domain. Moreover, for more challenging conditions where there are significant discrepancies in distributions of object sizes on the target domain (\ie, Nus$\rightarrow$KITTI), the statistical information for GCOS is necessary. How to reduce this dependence on statistical information worths further research.

\end{document}